\documentclass[letterpaper]{article} 
\usepackage{aaai2026}  
\usepackage{times}  
\usepackage{helvet}  
\usepackage{courier}  
\usepackage[hyphens]{url}  
\usepackage{graphicx} 
\usepackage{natbib}  
\usepackage{caption} 
\usepackage{algorithm}
\usepackage{algorithmic}

\usepackage{amsmath}
\usepackage{amssymb}
\usepackage{mathtools}
\usepackage{amsthm}
\usepackage{graphicx}
\usepackage{caption}
\usepackage{lipsum}
\usepackage[capitalize,noabbrev]{cleveref}
\usepackage{cite}
\usepackage{enumitem} 
\usepackage{url}            
\usepackage{latexsym}

\usepackage{booktabs}
\usepackage{threeparttable}

\usepackage{multicol}
\usepackage{multirow}
\usepackage{graphicx} 

\usepackage{pifont}

\usepackage{pifont}
\usepackage{mathrsfs} 
\usepackage{amsmath} 
\usepackage{colortbl} 
\usepackage{xcolor}   
\definecolor{lightblue}{RGB}{230, 242, 255}
\usepackage{bm}
\usepackage{makecell} 
\usepackage{newfloat}
\usepackage{listings}
\usepackage{tabularx}
\DeclareCaptionStyle{ruled}{labelfont=normalfont,labelsep=colon,strut=off} 
\floatstyle{ruled}
\newfloat{listing}{tb}{lst}{}
\floatname{listing}{Listing}
\title{TouchFormer: A Robust Transformer-based Framework for Multimodal Material Perception}

\author{
Kailin Lyu\textsuperscript{\rm 1,2},
Long Xiao\textsuperscript{\rm 1,2},
Jianing Zeng\textsuperscript{\rm 1}, 
Junhao Dong\textsuperscript{\rm 3},
Xuexin Liu\textsuperscript{\rm 1},
Zhuojun Zou\textsuperscript{\rm 1},
Haoyue Yang\textsuperscript{\rm 1}, 
Lin Shu\textsuperscript{\rm 1}, 
Jie Hao\thanks{Corresponding author.}\textsuperscript{\rm 1}
}

\affiliations{
\textsuperscript{\rm 1}Institute of Automation, Chinese Academy of Sciences\\
\textsuperscript{\rm 2}School of Artificial Intelligence, University of Chinese Academy of Sciences\\
\textsuperscript{\rm 3}Nanyang Technological University\\
}

\begin{document}

\maketitle

\begin{abstract} 


Traditional vision-based methods for material perception often experience substantial performance degradation under visually impaired conditions, thereby motivating the shift toward non-visual multimodal material perception. However, current approaches typically fuse modalities naively, overlooking critical challenges including modality-specific noise, the frequent absence of modalities, and their dynamically varying importance across tasks. These limitations lead to suboptimal performance across several benchmark tasks. In this paper, we propose a robust multimodal fusion framework, TouchFormer.
Specifically, we employ a Modality-Adaptive Gating (MAG) mechanism and intra- and inter-modality attention mechanisms to adaptively integrate cross-modal features, enhancing model robustness. We further introduce a Cross-Instance Embedding Regularization (CER) strategy to enhance performance in fine-grained subcategory material recognition tasks. Experimental results demonstrate that, compared to existing non-visual methods, the proposed TouchFormer framework achieves classification accuracy improvements of 2.48\% and 6.83\% on SSMC and USMC tasks, respectively. Additionally, real-world robotic experiments validate TouchFormer's effectiveness in enabling robots to better perceive and interpret their environment, paving the way for its deployment in safety-critical applications such as emergency response and industrial automation. 

\end{abstract}

\begin{links}
    \link{Website}{https://touchformer.github.io/TouchFormer/}
\end{links}

\section{Introduction}
\label{Introduction}


Material perception is a critical capability for both humans and robots when interacting with objects, typically relying on multiple modalities such as vision and touch~\citep{komatsu2018neural,material2}. Seen Surface Material Classification  (SSMC)~\citep{liu2023surface,khojasteh2024multimodal} and Unknown Surface Material Classification  (USMC)~\citep{wangsuo} are currently the two most representative material perception tasks, distinguished by whether the task involves recognizing previously unseen categories. Previous work has achieved good performance in scenarios where vision is normal or partially constrained~\citep{tatiya2024mosaic,khojasteh2024multimodal,song2025reconvla}. However, in scenarios where vision is completely unavailable, such as fire scenes, foggy conditions, or dark factories, the performance of vision-based methods may be affected or even significantly degraded. Therefore, non-visual material perception becomes particularly important. 

\begin{figure}[!t]
    \centering 
    \includegraphics[width=\linewidth]{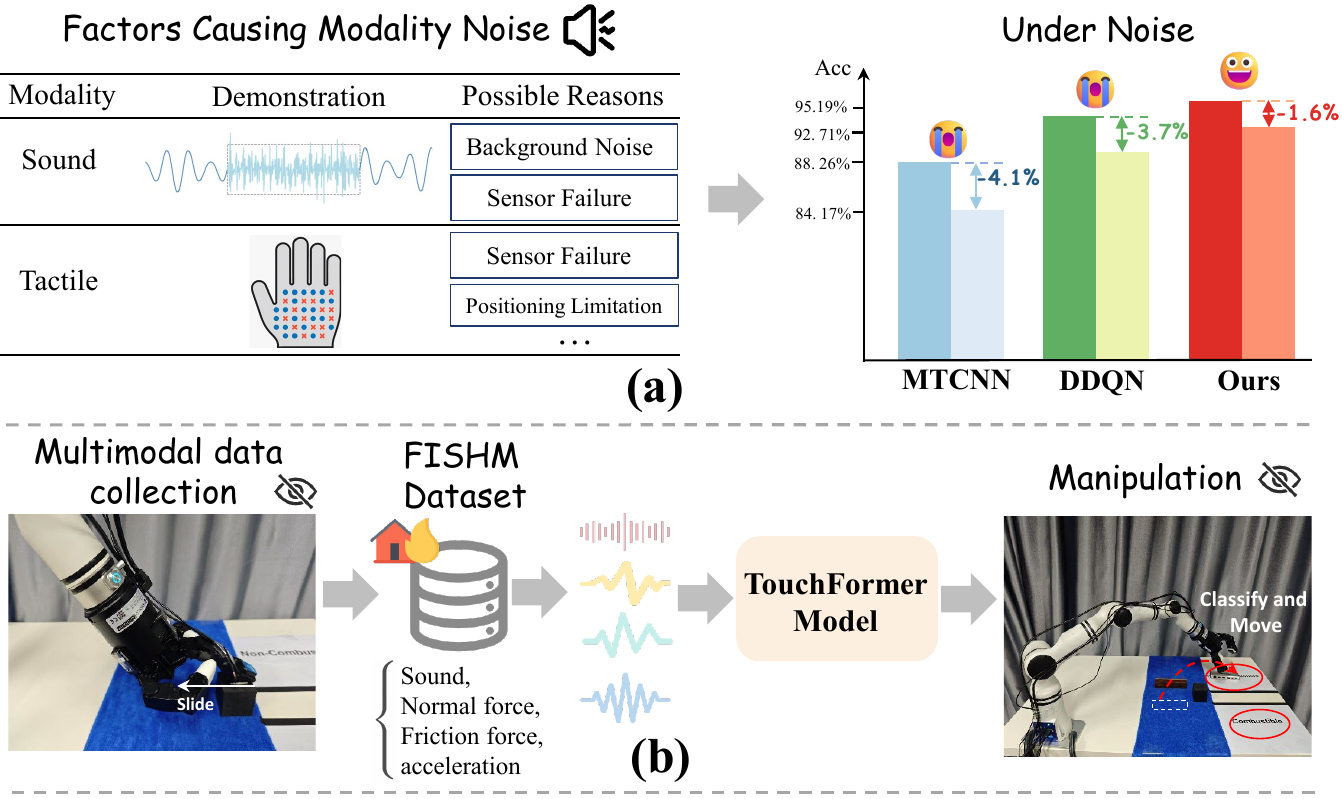} 
    \caption{\textbf{(a)} In the SSMC task, when facing real-world modality noise, TouchFormer shows only minor performance loss, in contrast to substantial degradation in baselines like MTCNN~\citep{wangsuo} and DDQN~\citep{liu2023surface}. \textbf{(b)} Under the simulated fire scenario prototype, the TouchFormer model collects and processes multimodal data (FISHM). Then, leveraging a physics engine to enable robust material perception and manipulation under extreme and vision-constrained conditions.}
    \label{fig:intro} 
\end{figure}

Although a small number of studies have begun to incorporate multiple non-visual modalities, such as touch and hearing, into material recognition~\citep{liu2023surface,wangsuo}, they often overlook the \textbf{non-ideality} of data in real-world scenarios. First, sensors across different modalities frequently operate at varied sampling rates, resulting in inherent temporal misalignment between modalities~\citep{MuIT}. Second, acquiring multimodal data in real-world scenarios inevitably involves noise or sensor failures, which can contaminate or omit information received by the model. This can directly lead to a significant drop in the performance of existing models(Figure 1(a)). Therefore, ensuring the model’s robustness to input data in real-world scenarios is essential for handling cases where multimodal inputs may be temporally misaligned, partially corrupted, or incomplete~\citep{junhao3}. Moreover, previous multimodal fusion algorithms typically assign equal weights to all modalities and fuse them directly~\citep{wangsuo}. However, this is often unreasonable in real-world scenarios. The key modality for identifying different materials often varies~\citep{adaptive1,adaptive2}, and assuming equal weights for all modalities may weaken the advantage of the critical modality, thus reducing the general accuracy of the model.

In this paper, we propose a robust multimodal fusion framework, called TouchFormer, designed to address two major issues in existing models: insufficient robustness to input data and suboptimal modality fusion strategies. Specifically, TouchFormer takes noisy or incomplete multimodal sequences as input and employs a Modality-Adaptive Gating (\textbf{MAG}) mechanism to dynamically assess the quality of each modality and assign appropriate weights. It then integrates multiple temporally misaligned input modalities through intra-modal and inter-modal attention mechanisms for adaptive fusion based on their importance. As a result, it can effectively extract relevant information from imperfect data to produce enhanced fused modality representations. Building on this, we introduce a Cross-Instance Embedding Regularization (\textbf{CER}) strategy to further improve the representational capacity of the embedding features. In conclusion, our main contributions are summarized as follows:

\begin{enumerate}
    \item To address the issues of insufficient robustness to input data and suboptimal modality fusion in existing non-visual multimodal models, we propose \textbf{TouchFormer}, a robust multimodal representation learning framework.
    \item We propose three functionally complementary core modules: Modality-Adaptive Gating (MAG), intra- and inter-modal attention mechanisms, and Cross-Instance Embedding Regularization (CER), which together enhance the robustness and representational capacity of TouchFormer.
    \item Experimental results demonstrate that TouchFormer outperforms existing non-visual multimodal approaches, achieving accuracy gains of at least \textbf{2.48\%} on the SSMC task and \textbf{6.83\%} on the USMC task. In the analysis of robustness to noisy modalities, TouchFormer also achieves the best robustness among existing non-visual multimodal models.
    \item To evaluate the model's performance in complex real-world scenarios, we proposed the Fire Incident Sound and Haptic Material~\textbf{(FISHM)} dataset and simulated a robotic material-sorting task under blindfolded fire conditions. This further validates the effectiveness of the TouchFormer framework in enabling robots to identify materials without relying on vision (Figure 1(b)), demonstrating its promising potential for applications in emergency response and industrial settings.
\end{enumerate}

\section{Related Works}
\label{Related Works}



\subsection{Non-visual Material Perception}
\label{Non-visual Material Perception}  

In visually impaired environments, touch is the primary perception modality, with robots detecting information such as force and acceleration to sense objects~\citep{calandra2018more,yuan2018active,sunil2023visuotactile}. Additionally, a few studies have utilized audition to distinguish material properties~\citep{shan2025star,tatiya2024mosaic}. Existing work typically integrates multiple modalities (e.g., normal force, friction force, and audition) to leverage their complementarity and enhance perception accuracy~\citep{bhattacharjee2018multimodal,liu2023surface}. Wei et al. introduced MTCNN, which integrates energy-spectrum features, dilated convolutions, and sequence pooling into a unified multimodal temporal convolutional network, enabling efficient fusion of acoustic and tactile cues for material recognition~\citep{wangsuo}. hojasteh et al. proposed MMUSR, a data-versus-data framework based on the kernel two-sample test that performs material classification on heterogeneous data with minimal manual tuning~\citep{khojasteh2024multimodal}. However, existing methods merely perform simple fusion of multiple modalities and require temporal alignment, resulting in poor robustness in real-world scenarios. Our proposed framework allows for input data to be corrupted or partially missing, demonstrating strong potential in real robotic tasks.

\subsection{Tactile Sensors}
\label{Types of Tactile Sensors}

In recent years, tactile sensors have been widely adopted in various robotic applications, including slip detection, object manipulation, insertion, and material recognition~\citep{james2020slip, dahiya2013directions}. These sensors can generally be classified into two main categories. The first category comprises vision-based tactile sensors (VBTS)~\citep{dong2017improved,lambeta2020digit}, which capture detailed information about object shape and material properties by observing the deformation of an illuminated membrane. Although these sensors provide high-resolution tactile data, they typically have a limited lifespan and are unsuitable for harsh environments such as fire or disaster scenarios. The second category, represented by uSkin~\citep{uskin}, consists of multi-contact tactile sensors that measure multiple types of signals, including force, vibration, and acceleration, through simple and low-dimensional sensing mechanisms. These sensors are known for their low cost and strong durability~\citep{paulino2017low,tomo2016modular}. Considering the demanding requirements for robustness and reliability in field environments, this study employs the uSkin sensor, which is capable of simultaneously measuring both normal and frictional forces.

\begin{figure*}[htbp] 
    \centering 
    \includegraphics[width=\textwidth]{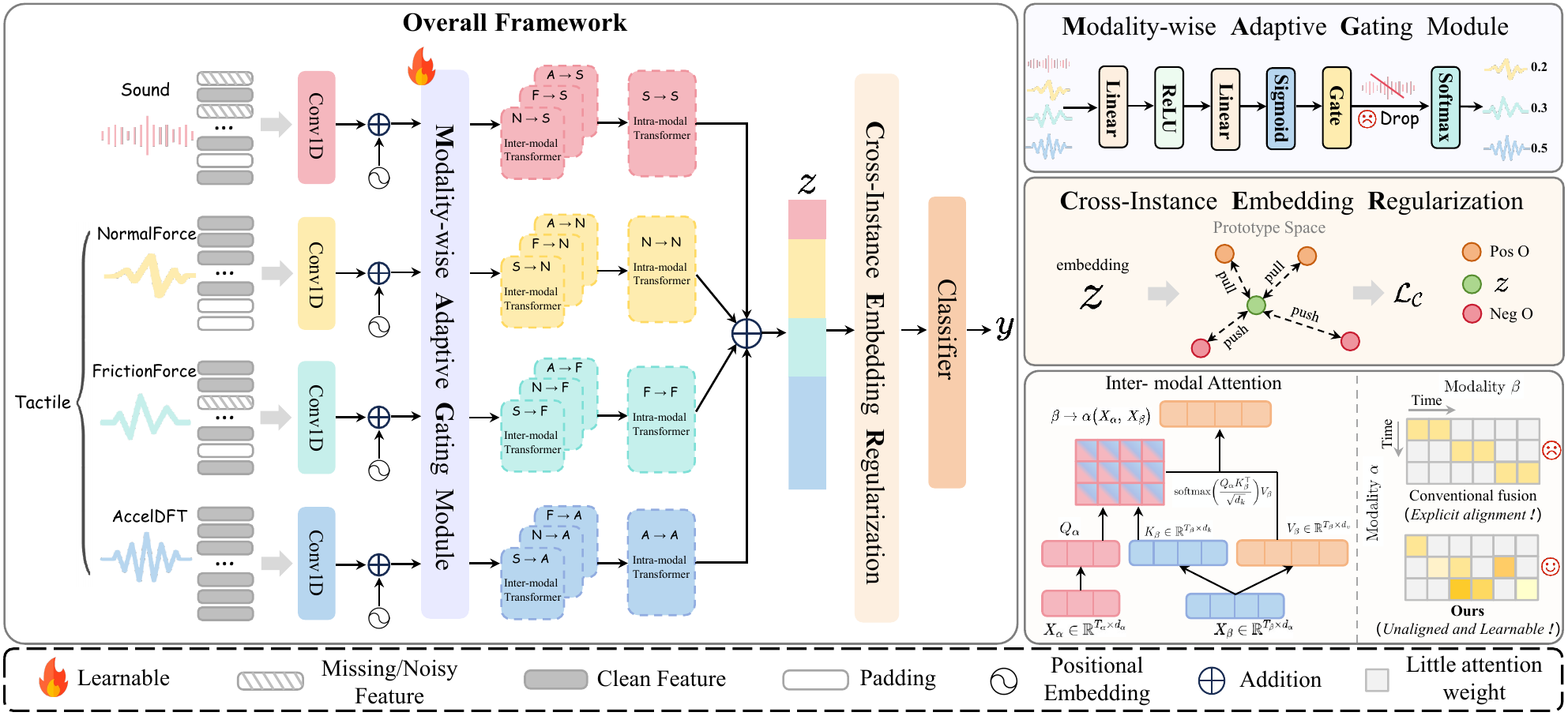}
    \caption{\textbf{Overview of the proposed framework.} TouchFormer receives noisy or incomplete multimodal sequences as input and employs Modality-Adaptive Gating~\textbf{(MAG)} to dynamically assess the quality of each modality. It then adaptively integrates cross-modal features within the latent space using both intra- and inter-modality attention mechanisms, without requiring explicit alignment. Finally, Cross-Instance Embedding Regularization~\textbf{(CER)} is applied to improve the clarity and discriminability of the representation space, thereby facilitating robust surface material classification.} 
    \label{fig:pipline} 
\end{figure*}

\section{Methodology}
\label{Methodology}


Figure~\ref{fig:pipline} illustrates the proposed TouchFormer framework, and the detailed pseudocode is illustrated in the \textbf{Appendix A}. It comprises three functionally complementary core modules: Modality-Adaptive Gating module~\textbf{(MAG)}, which enhances input reliability by filtering noisy or irrelevant modalities at the source through dynamic weighting. \textbf{Intra- and inter-modal} Transformer fusion module facilitates deep integration across modalities, addressing asynchronous alignment and cross-modal interaction, and cross-instance embedding regularization module~\textbf{(CER)} further optimizes the prototype space to enhance feature discriminability. Together, the three modules form a complementary system that enhances \textbf{robustness}, \textbf{fusion quality}, and \textbf{discriminability}, enabling reliable Multimodal material perception in complex real-world environments.

\subsection{Modality-Adaptive Gating Module}
\label{Modality-Adaptive Gating Module}

Multimodal data are inherently noisy and often incomplete, leading to significant disparities in the quality of information provided by different modalities. However, traditional multimodal fusion methods~\citep{khojasteh2024multimodal} typically treat features from different modalities equally, such as by assigning uniform weights or directly concatenating them, without considering modality-specific reliability. This uniform strategy fails to handle biased or noisy inputs, potentially causing negative transfer and performance degradation (Section~\ref{Introduction}). To address this issue, we propose a \textbf{MAG} module, which dynamically evaluates and adaptively adjusts the importance of each modality during feature fusion. Specifically, for each modality $X_{m}\in\mathbb{R}^{T\times d}$, an intermediate feature representation $H_{m}$ is first computed via a linear transformation followed by a nonlinear activation function:
\begin{equation}
H_{m} = \text{ReLU}(W_{1}X_{m}+b_{1}).
\end{equation}

Subsequently, modality-specific gating weights $g_{m}\in[0,1]$ are generated by applying another linear transformation and a sigmoid activation function to the intermediate representation:
\begin{equation}
g_{m} = \sigma(W_{2}H_{m}+b_{2}).
\label{eq2}
\end{equation}

We further introduce a hyperparameter $gate_{th}$. Modalities with gating weights below this threshold ($g_{m}<gate_{th}$) are considered as providing insufficient or noisy information and thus discarded to prevent contamination of the fused multimodal representations.

Finally, to explicitly quantify the relative contributions of each modality(sound \(S\), normal force \(N\), friction force \(F\), and acceleration \(A\)) during the fusion process, we apply a softmax normalization on the gating weights to obtain the final modality importance weights $\alpha_{m}$:
\begin{equation}
\alpha_{m} = \frac{\exp(g_{m})}{\sum_{k}\exp(g_{k})}, \quad m \in \{S, N, F, A\}.
\end{equation}

The modality features adjusted through the aforementioned adaptive gating are computed as follows:
\begin{equation}
Z_{m} = \alpha_{m} \odot (X_{m} + PE(T,d)),
\label{eq4}
\end{equation}
where $PE(T,d)$ denotes positional embeddings, and $\odot$ denotes the element-wise multiplication operation. This operation enables the model to dynamically adapt to different modalities, filtering out low-quality modalities and improving input reliability, thereby enhancing the robustness of subsequent feature fusion.

\subsection{Inter- and Intra-modal Transformer Fusion Module}
\label{Intra- and Inter-modal Transformer Fusion Module} 

Conventional multimodal fusion adopts two paradigms: \textbf{\textit{i)}} concatenating raw or intermediate features at a fixed layer before a Transformer~\citep{concate1,concate2}, and \textbf{\textit{ii)}} using convolutional networks to extract modality specific features and then concatenating them~\cite{confuse1,liu2024maniwav}. Both strategies generally require manual alignment of modality sequences to a common time step. In embodied-intelligence perception scenarios, however, heterogeneous \textbf{sensors exhibit inherent latency}, making such alignment both \textbf{labor-intensive} and \textbf{error-prone}. To capture within-modality temporal structures and cross-modality semantic dependencies, we simultaneously model intra- and inter-modal interactions. Inspired by MulT (Tsai et al., 2019), we adopt cross-modal attention to inject low-level source features into target modalities through explicit cross modal attention, enabling robust fusion of asynchronous sequences without alignment. Unlike MulT, our method applies \textbf{MAG} again before final feature integration, reweighting block negative transfer, enabling further adaptive fusion and enhancing both flexibility and robustness. 

\noindent\textbf{Temporal Convolution and Positional Embedding.} To allow each sequence element to perceive its local neighbourhood, we first apply a one–dimensional temporal convolution to the four modalities \(X_{m}\):
\begin{equation}
\hat{X}_{m}= \operatorname{Conv1D}\!\bigl(X_{m},k_{m}\bigr)
\in \mathbb{R}^{T_{m}\times d},\quad
m\in\{S,N,F,A\},
\end{equation}
where \(k_{m}\) denotes the kernel size of modality \(m\) and \(d\) is the unified feature dimension.  
We then add positional embeddings:
\begin{equation}
Z_{m}^{[0]}=\hat{X}_{m}+PE(T_{m},d).
\end{equation}

\noindent\textbf{Inter-modal Transformer.} Let $\alpha$ be the target modality and $\beta$ the source modality.
We first compute the standard cross-modal attention
\begin{equation}
\hat{Y}_{\alpha\leftarrow\beta}=%
\operatorname{softmax}\!\Bigl(
  \tfrac{Q_{\alpha}K_{\beta}^{\top}}{\sqrt{d_{k}}}
\Bigr)V_{\beta},
\label{eq7}
\end{equation}

The attention output is subsequently modulated by the modality importance weight \(\alpha_{\beta}\), which is computed by the \textbf{MAG} module (Section~3.1), such that \(Y_{\alpha \leftarrow \beta} = \alpha_{\beta}\, \hat{Y}_{\alpha \leftarrow \beta}\). \(\alpha_{\beta} \in [0, 1]\) reflects the reliability of the source modality \(\beta\); a higher weight indicates a greater contribution to the representation of the target modality.

\noindent\textbf{Intra-modal Transformer.} The inter-modal output is merged with the original representation by a residual connection:
\begin{equation}
\tilde{Z}_{\alpha}=Z_{\alpha}^{[0]}+Y_{\alpha\leftarrow\beta}.
\end{equation}
Self-attention is then applied within the same modality:
\begin{equation}
Z_{\alpha}^{\mathrm{intra}}=
\operatorname{Transformer}\!\bigl(
\tilde{Z}_{\alpha},
\tilde{Z}_{\alpha},
\tilde{Z}_{\alpha}
\bigr).
\label{eq11}
\end{equation}
Finally, the intra-modal representations of the four modalities,
\(Z_{m}^{\mathrm{intra}}\), are re-weighted by their importance coefficients
\(\alpha_{m}\) and concatenated to form the fused feature vector:
\begin{equation}
Z=
\operatorname{Concat}\!\bigl[
\alpha_{S}Z_{S}^{\mathrm{intra}},
\alpha_{N}Z_{N}^{\mathrm{intra}},
\alpha_{F}Z_{F}^{\mathrm{intra}},
\alpha_{A}Z_{A}^{\mathrm{intra}}
\bigr].
\label{eq12}
\end{equation}

This sequence jointly captures intra-modal temporal dynamics and inter-modal interactions, enabling the model to learn rich, complementary features for robust multimodal representation.

\subsection{Cross-Instance Embedding Regularization}
\label{Cross-instance Embedding Regularization Module}
In multimodal representation learning, despite improved information integration through modality selection and fusion strategies, models may still exhibit inter-class confusion and intra-class scattering~\citep{zhou2023exploiting,jiang2023understanding}. To address these limitations and enhance both the discriminative capacity and generalization of learned representations, we introduce a Cross-Instance Embedding Regularization~\textbf{(CER)} module. Grounded in contrastive learning principles, CER enforces structural constraints on the embedding space from a cross-instance perspective by leveraging global supervision signals. It promotes intra-class compactness and inter-class separability, thereby enhancing the clarity and discriminability of the overall representation space.

Given a batch of $N$ samples with $\ell_{2}$-normalized embeddings $\{z_{i}\}_{i=1}^{N}$ and corresponding labels $\{y_{i}\}_{i=1}^{N}$, we construct the similarity matrix $S_{ij}=z_{i}^{\top}z_{j}$. The cross-instance contrastive loss
$\mathcal{L}_{C}$ is defined as
\begin{equation}
\mathcal{L}_{C}= -\frac{1}{N}\sum_{i=1}^{N}
\log
\frac{\displaystyle
      \sum_{\,j\neq i}\mathbf{I}_{\{y_{i}=y_{j}\}}\,
      \exp\!\bigl(S_{ij}/\tau\bigr)}
     {\displaystyle
      \sum_{\,j\neq i}\exp\!\bigl(S_{ij}/\tau\bigr)},
\label{eq:contrastive}
\end{equation}
where $\tau$ is the temperature hyper-parameter and $\mathbf{I}_{\{y_{i}=y_{j}\}}$ is an indicator function that equals~1 only for positive pairs $(y_{i}=y_{j})$. This loss pulls together embeddings from the same class while pushing apart those from different classes, encouraging a discriminative feature space. As it operates on relative distances rather than fixed labels, it is well suited for tasks requiring instance-level semantic mapping (Section~\ref{Fine-Grained Subclass Classification}).

\subsection{Overall Loss}
\label{Model Training} 

The model is trained in a single stage by jointly minimizing the classification loss
$\mathcal{L}_{cls}$ and the CER loss
$\mathcal{L}_{C}$:
\begin{equation}
\mathcal{L}_{total} = \mathcal{L}_{cls}(y,\hat{y}) + \lambda\,\mathcal{L}_{C},
\label{eq14}
\end{equation}
where \(\hat{y}\) denotes the predicted labels, \(y\) the ground-truth
labels, and \(\lambda\) is a weighting factor that controls the influence
of $\mathcal{L}_{total}$ on the total objective. The parameters of the \textbf{MAG} module are updated concurrently with the network. Minimizing $\mathcal{L}_{C}$ enhances both the robustness of the multimodal representations and the model’s ability to generalize in classification tasks.

\section{Experiment Setup}
\label{Experiment Setup}

\subsection{Datasets}
\label{Datasets}

\textbf{LMT Haptic Material Database (LMTHM)}: This publicly available multimodal dataset was collected by Strese et al.~\citep{LMT} using the self-developed Texplorer2 device. The dataset comprises 965 samples from 193 distinct surface materials, with five samples per material. It includes multimodal data such as sound, acceleration, normal force, and frictional force. The materials are categorized into eight major classes and several subclasses.

\begin{figure}[!tbp]
    \centering 
    \includegraphics[width=0.85\linewidth]{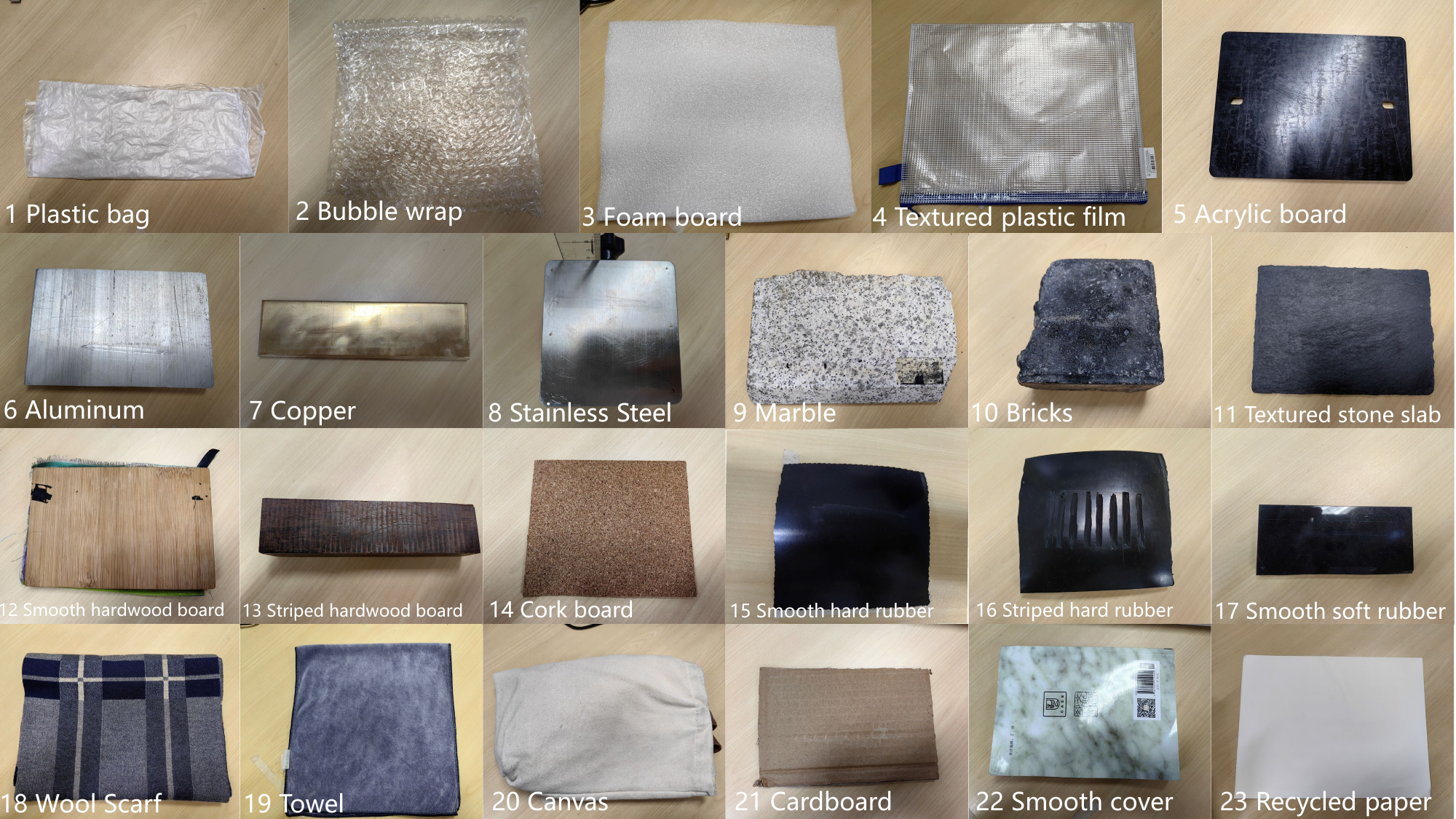} 
    \caption{The FISHM dataset comprises seven distinct categories of daily objects, encompassing representative item types potentially encountered in fire incidents.} 
    \label{fig:FISHM} 
\end{figure}

\noindent
\textbf{Fire Incident Sound and Haptic Material (FISHM) Dataset} Figure~\ref{fig:FISHM}, acquired through a custom multimodal tactile fingertip. The integrated fingertip employs a uSkin tactile sensor for Normal and Frictional force measurement, a microphone for Sound measurement, and a 12-DoF IMU for tri-axial Acceleration measurement, enabling simultaneous collection of four modalities. It comprises 7 daily material categories, with metal and stone classified as \textit{Non-combustible} and others as \textit{Combustible}. Furthermore, it accurately simulates fire scenarios, thereby facilitating a more effective evaluation of the robustness and generalizability of the proposed model. Comprehensive details regarding the custom figertip, FISHM dataset and multimodal data collection process are provided in \textbf{Appendices B and C}. 


\subsection{Baselines}
\label{Baselines}

We carefully selected widely used state-of-the-art robotic sensing techniques based on the uSkin tactile sensor for multimodal surface material classification tasks to compare with our proposed method, TouchFormer. Specifically, for methods \textbf{excluding the visual modality}, we selected LSTM~\citep{ji2015preprocessing}, MTCNN~\citep{wangsuo}, and DDQN~\citep{liu2023surface} as baseline models, each capable of classifying surface materials solely using auditory and multiple tactile modalities (including acceleration, normal force, and frictional force). For methods \textbf{incorporating the visual modality}, we chose the recently proposed MMUSR~\citep{khojasteh2024multimodal}, which uses up to nine different sensor modalities (e.g. visual and tactile signals) for classification.

\subsection{Implementation Details}
\label{Implementation Details}

All experiments were conducted on machines equipped with A100 GPUs using the PyTorch framework. During training, we used a batch size of 32 and the Adam optimizer with a weight decay of 0.1. The initial learning rate was set to 0.1 and gradually decayed to 0 using a cosine annealing strategy~\citep{loshchilov2016sgdr}. The models were trained for a total of 50 epochs. In the robotic application phase, the model trained on the LMTHM dataset was deployed on a robotic arm and fine-tuned using the FISHM dataset for domain adaptation. For the USMC task, we adopted a fivefold cross-validation strategy, consistent with the evaluation protocol used by MTCNN~\citep{wangsuo}, to ensure a fair comparison across methods. For the SSMC task, we split the dataset into training and testing sets at a 7:3 ratio. In both settings, performance was evaluated using the mean classification accuracy and G\_mean~\citep{gmean} as the primary metrics.

\section{Experimental Results}
\label{Experimental Results}

In this section, we conduct experiments based on the dataset collected using the uSkin sensor to evaluate the performance of our model in surface material classification under various conditions. The objectives are as follows: \textbf{\textit{i).}} To verify whether the proposed TouchFormer framework outperforms baseline methods and to demonstrate the effectiveness of each component within the TouchFormer framework. \textbf{\textit{ii).}} To assess the model’s robustness under randomly corrupted modality features. \textbf{\textit{iii).}} To evaluate the framework’s performance in enabling robotic environmental understanding in real-world physical scenarios.

\begin{table}[htbp]
\centering
\small
    \begin{tabular}{ccccc}
    \toprule
    Task & \makecell{Class \\ num}  & {Method} & {Accuracy [\%]} & {G\_mean} \\

    \midrule
    \multirow{4}{*}{SSMC} & \multirow{4}{*}{8}
    & DDQN     & 92.71 & 0.93  \\
    & & MMUSR & 79.7  & 0.80  \\
    & & MMUSR$^\star$ & 93.8  & 0.94  \\
    & & \cellcolor{lightblue}\textbf{Ours} & \cellcolor{lightblue}\textbf{95.19} (↑ 2.48) & \cellcolor{lightblue}\textbf{0.95}  \\

    \hline\hline

    \multirow{4}{*}{USMC} & \multirow{4}{*}{8}
    & LSTM\textsuperscript{\textdagger} & 74.23 & 0.73 \\
    & & DALNet\textsuperscript{\textdagger}   & 46.67 & 0.47  \\
    & & MTCNN\textsuperscript{\textdagger}           & 87.55 & 0.88  \\
    & & \cellcolor{lightblue}\textbf{Ours} & \cellcolor{lightblue}\textbf{94.38} (↑ 6.83) & \cellcolor{lightblue}\textbf{0.94}   \\

    \hline\hline

    \multirow{4}{*}{\makecell{SSMC\\(Fine-\\Grained)}} & \multirow{4}{*}{193} 
    & LSTM  & 65.75 & 0.66  \\
    & & DALNet   & 42.62 & 0.43  \\
    & & MTCNN           & 80.21 & 0.80  \\
    & & \cellcolor{lightblue}\textbf{Ours} & \cellcolor{lightblue}\textbf{89.77} (↑ 9.56) & \cellcolor{lightblue}\textbf{0.90}  \\
    \bottomrule
    \end{tabular}
\normalsize
\caption{\textbf{Multimodal surface material classification} performance on the LMTHM dataset. "$\star$" indicates that the model employs visual input. "\textdagger" denote the corresponding sources of the reported results~\citep{wangsuo}.}
\label{tab: main result}
\end{table}




\begin{table}[htbp]
\centering
\begin{small}
    \begin{tabular}{ccccccc}
    \toprule
    \multicolumn{4}{c}{Multimodal Inputs} & \multirow{2}{*}{LSTM}& \multirow{2}{*}{MTCNN} & \multicolumn{1}{c}{\multirow{2}{*}{\textbf{Ours}}} \\
    \cmidrule(lr){1-4}
    $\boldsymbol{S}$ & $\boldsymbol{N}$ & $\boldsymbol{F}$ & $\boldsymbol{A}$ & & \\
    \midrule
    \ding{55} & \ding{51}& \ding{51}& \ding{51} & 73.78\% & 83.43\% & \textbf{87.92\%}\\
    \ding{51} & \ding{55}& \ding{51}& \ding{51} & 71.69\% & 83.18\% & \textbf{88.55\%}\\
    \ding{51} & \ding{51}& \ding{55}& \ding{51} & 69.09\% & 84.99\%& \textbf{89.12\%}\\
    \ding{51} & \ding{51}& \ding{51}& \ding{55} & 75.50\% & 87.00\%& \textbf{89.84\%}\\
    \ding{51} & \ding{51}& \ding{51}& \ding{51} & 74.23\% & 87.55\%& \textbf{94.38\%}\\
    \bottomrule
    \end{tabular}
\end{small}

\caption{\textbf{Comparison of classification accuracy} as sound (S), normal force (N), friction force (F), and acceleration (A) modalities are incrementally removed. Even with any single modality missing, TouchFormer consistently outperforms other methods using full-modal inputs.}
\label{tab: main result2}
\end{table}

\subsection{Multimodal Performance Comparison}
\label{Multimodal Performance Comparison}

We investigated whether our proposed method outperforms existing benchmark approaches under multimodal conditions and assessed the necessity and effectiveness of integrating auditory and multi-tactile data for the surface material classification task. The results presented in Table~\ref{tab: main result} demonstrate that, in multimodal scenarios, the TouchFormer framework achieves classification performance improvements of 2.48\% and 6.83\% over the best baseline methods on the SSMC and USMC tasks, respectively. Notably, even with no visual modality input, our method still surpasses the MMUSR~\citep{khojasteh2024multimodal} includes visual modality input and achieves an accuracy close to the 97.2\% performance obtained by MMUSR, which integrates a total of eight modalities derived from six different sensors, including a camera, multiple tactile sensors, an infrared(IR) surface reflectance sensor and a metal detection sensor. Additionally, Table~\ref{tab: main result2} shows that in the USMC task, even after individually removing modalities such as sound, normal force, friction force, and acceleration, TouchFormer still outperforms other methods.

\subsection{Fine-Grained Subclass Classification}
\label{Fine-Grained Subclass Classification}

\begin{figure}[htbp]
    \centering 
    \includegraphics[width=0.95\linewidth]{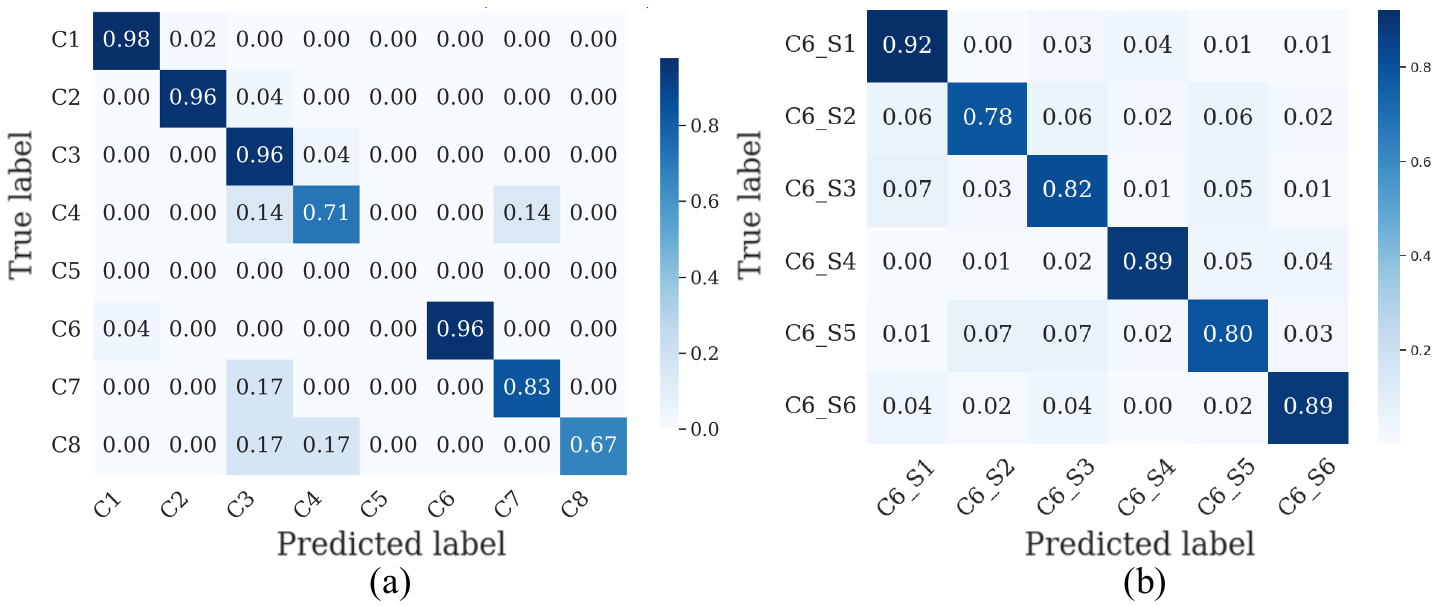} 
    \caption{TouchFormer confusion matrix performance on the SSMC task. (a) Coarse-grained classification results across 8 material classes. (b) Fine-grained classification results (e.g., using category C6 as an example).} 
    \label{fig:matrix} 
\end{figure}

\begin{figure}[htbp]
    \centering 
    \includegraphics[width=0.75\linewidth]{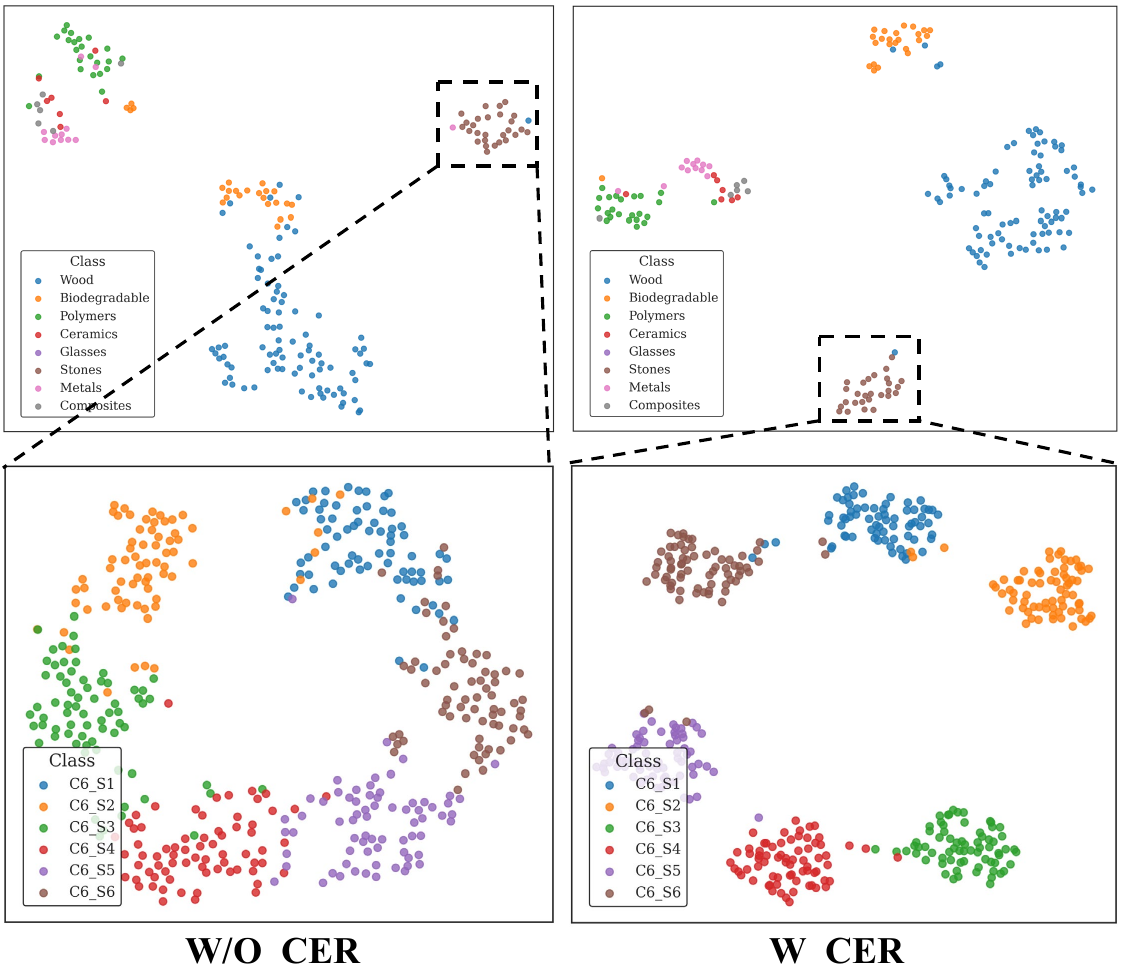}
    \caption{Visualization of subclass embeddings with and without CER. t-SNE plots of feature embeddings for coarse- and fine-grained classification. } 
    \label{fig:tsne} 
\end{figure}

Human tactile perception relies on both coarse-grained sensing and fine-grained discrimination primarily enabled by the fast-adapting (FA) and slow-adapting (SA) systems~\citep{park2016fast}. In addition to evaluating our method on a coarse-level surface material classification task with eight categories, we introduce a novel, more challenging fine-grained subclass classification task, such as distinguishing \textit{softwood} from \textit{hardwood} within the broader category of \textit{wood}, which previous research has not addressed. As shown in Table~\ref{tab: main result} and Figure~\ref{fig:matrix}, our proposed method achieves outstanding classification performance across all categories in this task. It is primarily attributed to the proposed MAG and CER, which improve classification accuracy by adaptively selecting and integrating the most relevant modalities and enhancing subclass discriminability within the prototype embedding space (Figure~\ref{fig:tsne}).


\subsection{Model Robustness for Modality Anomaly}
\label{Model Robustness to Modality Anomaly}

\begin{figure}[!t]
    \centering 
    \includegraphics[width=\linewidth]{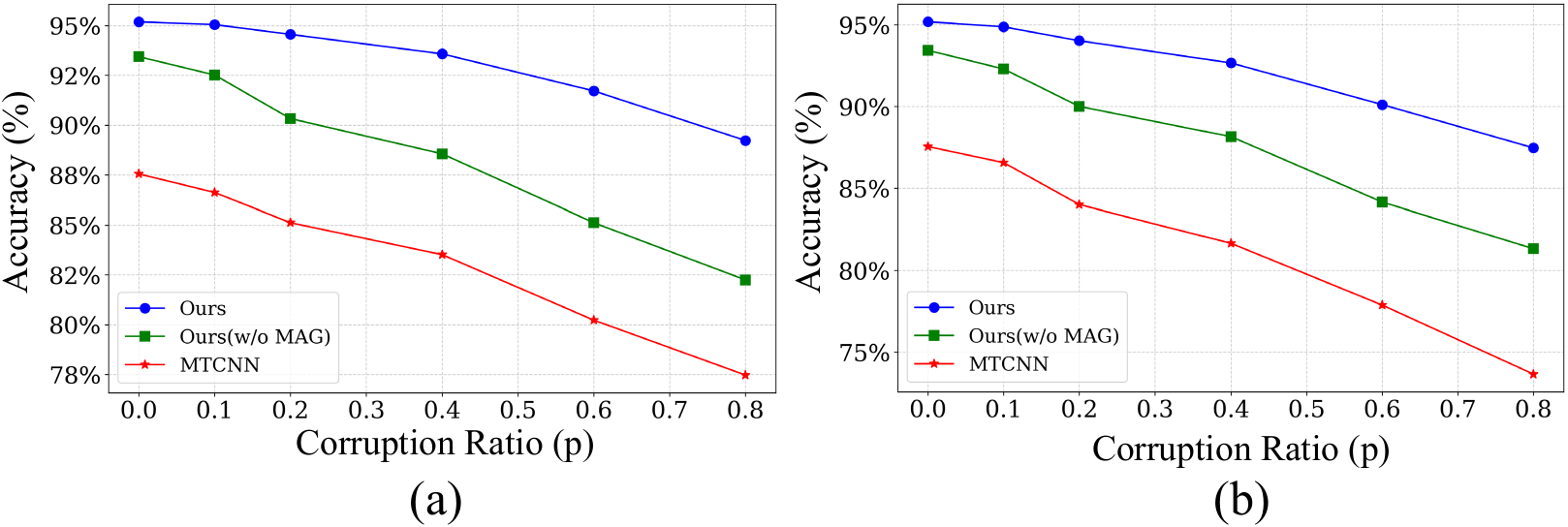} 

    \caption{\textbf{Performance under modality corruption.} We compare the classification performance of TouchFormer and other method under varying corruption ratios \( p \). } 
    \label{fig:plot curve} 
\end{figure}

We investigate the robustness of the TouchFormer framework under varying levels of noise applied to randomly selected modalities. Specifically, during both training and testing phases, we introduce noise to randomly selected modalities under different gaussian noise intensities, with a corruption ratio $p\in\{0.0, 0.1, \dots, 1.0\}$. As shown in Figure~\ref{fig:plot curve}, the TouchFormer framework consistently outperforms baseline methods across various noise combinations and intensity levels, demonstrating strong robustness in multimodal noisy environments. This improvement is primarily attributed to the \textbf{MAG} module, which dynamically evaluates and adjusts the contribution of each modality during feature fusion.

\begin{table}[!t]
\centering
\begin{small}
\begin{tabular}{cccc}
\toprule
Configurations & SSMC & USMC & Fine-Grained \\
\midrule
\rowcolor{gray!15}
Baseline & 91.32\% & 90.17\% & 83.53\% \\
Baseline + MAG & 93.15\% & 92.56\% & 84.88\% \\
Baseline + MAG + CER &\textbf{95.19\%}&\textbf{94.38\%}&\textbf{89.77\%} \\
\bottomrule
\end{tabular}
\end{small}
\caption{Ablation study of different components across diverse tasks}
\label{tab:ablation}
\end{table}



\begin{figure*}[htbp] 
    \centering 
    \includegraphics[width=\textwidth]{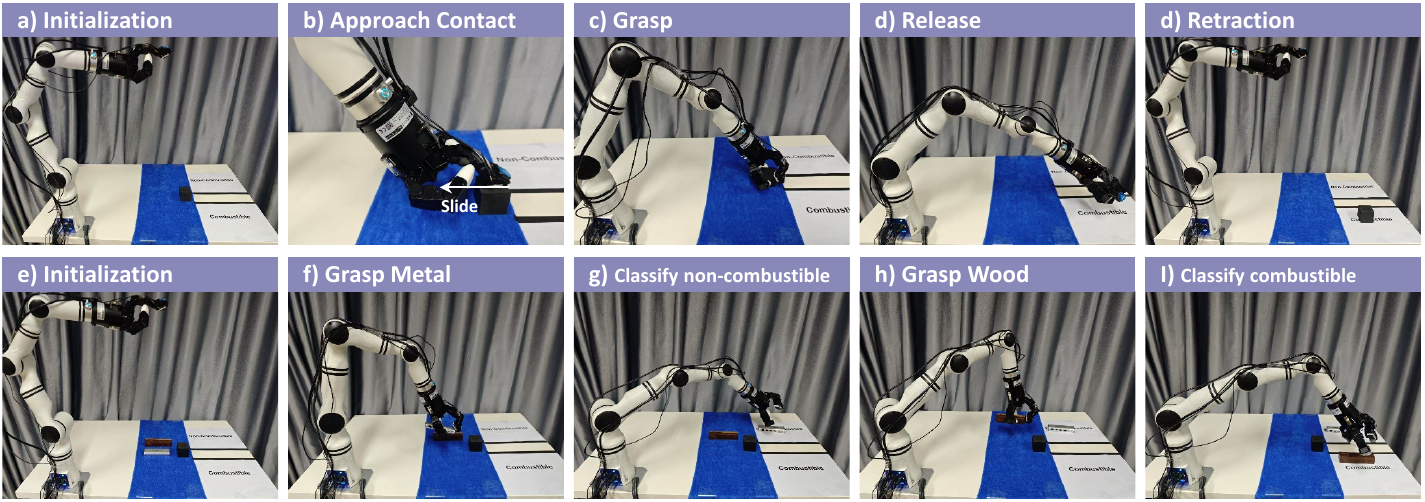}
    \caption{\textbf{Evaluation of material transport performance} using the RealMan robot in a vision-free, noise-disturbed simulated fire scenario. (a)$\sim$(d) illustrate the procedure for evaluating individual objects; (e)$\sim$(i) depict the evaluation process in multi-object scenes. The robot identifies the material properties of wood, rubber, and metal blocks, and executing appropriate grasp-and-place actions to move them into their corresponding zones.} 
    \label{fig:application} 
\end{figure*}

\subsection{Ablation Study}
\label{Ablation Study}



Table~\ref{tab:ablation} presents the ablation study results of the proposed TouchFormer model on the SSMC, USMC, and fine-grained classification tasks. The baseline refers to a configuration that employs only intra- and inter-modal Transformer fusion modules without any of the proposed enhancements. We first incorporate the MAG module into the baseline, which aims to adaptively select and fuse the most relevant modalities. This step improves performance from (91.32, 90.17, 83.53) to (93.15, 92.56, 84.88). Next, we introduce CER to enhance the discriminability of different categories in the prototype embedding space. After applying CER, classification accuracy improves by (+2.04\%, +1.82\%, +4.89\%). The corresponding visualization results are shown in Figures~\ref{fig:tsne} and~\ref{fig:matrix}. Additionally, we conduct further experiments by adjusting \(\mathit{gate}_{\mathit{th}}\) and \(\lambda\) to evaluate their impact on TouchFormer’s performance. While keeping all other settings consistent with Section~\ref{Implementation Details}, we test various threshold ranges and mixing ratios. The results are summarized in Appendix.


\subsection{Robotic Application}
\label{Robotic Application} 

Robotic material perception, such as surface material classification, aims to enhance robots' ability to understand and interact with their environments. While prior experiments validated our method’s effectiveness, a gap remains between perception and manipulation. To address this, we present a fire-scenario prototype demonstrating how material perception models guide strategy inference for robotic interaction.

We apply material perception techniques to a robotic sorting task in a simulated fire scenario, using a Realman RM65-B 6-DoF robotic arm equipped with a TESOLLO Gripper-3F three-fingered dexterous hand to perform object grasping and releasing operations. This task simulates key complexities inherent in fire environments, characterized by three principal constraints: compromised visibility, confined spatial configurations, and multiple objects. The procedure is as follows: (1) the robot system operates in a multi-object, low-visibility  environment where combustible and non-combustible objects remain fixed at predetermined positions, with designated collection zones on the lateral sides; (2) using multimodal perception, the robot identifies and classifies materials to determine their flammability; and (3) Based on the classification results, the robot executes a predefined program to move each object to its designated area while clearing the safety pathway. Since no visual input is available, object manipulation and grasping during the experiment are executed using fixed-parameter programs. The object category is determined solely by the algorithm. The setup is illustrated in Figure~\ref{fig:application}. We evaluate the performance of TouchFormer on various tasks using the FISHM dataset. The experimental results are summarized in Table~\ref{tab:FISHM_exp}. In addition, we assess the model’s real-world performance by integrating it with a physics engine. Specifically, we measure the classification accuracy of the robot system in identifying combustible and non-combustible materials within the designated collection zones during the SSMC task. Qualitative sorting process and results are illustrated in \textbf{Appendix D}.

\begin{table}[!t]
\centering
\begin{tabular}{cccc}
\toprule
Environments & SSMC & USMC &Fine-Grained \\
\midrule
w/o Noise & 91.47\% & 90.26\% & 90.05\% \\
w/  Noise & 89.54\% & 88.03\% & 87.76\% \\
\bottomrule
\end{tabular}
\caption{Performance of the TouchFormer on the FISHM dataset across different environments and tasks.}
\label{tab:FISHM_exp}
\end{table}

\section{Conclusion}
TouchFormer introduces a novel and robust framework for multimodal material perception under visually impaired and noisy conditions, achieving superior performance via adaptive and noise-aware fusion. The framework integrates a Modality-Adaptive Gating (MAG) mechanism alongside intra- and inter-modal Transformer-based fusion to explicitly account for modality noise and missing inputs, which are prevalent in real-world settings. Furthermore, the proposed Cross-Instance Embedding Regularization (CER) strategy enhances feature discriminability, particularly for fine-grained material classification. Extensive experiments demonstrate that TouchFormer consistently outperforms existing vision-free baselines across multiple benchmarks. Additionally, real-world robotic experiments confirm its capability to support accurate environmental understanding and informed action inference, underscoring its strong potential for deployment in high-stakes scenarios such as emergency response and industrial automation.

\section*{Acknowledgments}
This work was supported by the National Natural Science Foundation of China under Grant 62236007.

\bibliography{aaai2026}

\end{document}